\documentclass[sigconf, nonacm]{acmart}

\AtBeginDocument{%
  }

\copyrightyear{2023}
\acmYear{2023}
\setcopyright{acmlicensed}\acmConference[CODAI '23 ]{Workshop on
Compilers, Deployment, and Tooling for Edge AI}{September 21,
2023}{Hamburg, Germany}
\acmBooktitle{Workshop on Compilers, Deployment, and Tooling for Edge AI
(CODAI '23 ), September 21, 2023, Hamburg, Germany}
\acmPrice{15.00}
\acmDOI{10.1145/3615338.3618122}
\acmISBN{979-8-4007-0337-9/23/09}

\usepackage{textcomp}
\begin{document}

\title{Scaling Up Quantization-Aware Neural Architecture Search for Efficient Deep Learning on the Edge}

\author{Yao Lu}
\authornote{Equal contribution.}
\email{yao.lu_2@nxp.com}
\orcid{0009-0005-1614-6025}
\affiliation{%
  \institution{NXP Semiconductors}
  \city{Munich}
  \country{Germany}
}

\author{Hiram Rayo Torres Rodriguez}
\orcid{0000-0003-1202-4854}
\authornotemark[1]
\email{hiram.rayotorresrodriguez@nxp.com}
\affiliation{%
  \institution{NXP Semiconductors}
  \city{Eindhoven}
  \country{The Netherlands}
}

\author{Sebastian Vogel}
\orcid{0000-0001-9665-6562}
\affiliation{%
  \institution{NXP Semiconductors}
  \city{Munich}
  \country{Germany}}

\author{Nick van de Waterlaat}
\orcid{0000-0002-4577-6481}
\affiliation{%
  \institution{NXP Semiconductors}
  \city{Eindhoven}
  \country{The Netherlands}
}

\author{Pavol Jancura}
\orcid{0000-0002-9219-5014}
\affiliation{%
 \institution{Eindhoven University of Technology}
   \city{Eindhoven}
  \country{The Netherlands}}


\begin{abstract}
Neural Architecture Search (NAS) has become the de-facto approach for designing accurate and efficient networks for edge devices. Since models are typically quantized for edge deployment, recent work has investigated quantization-aware NAS (QA-NAS) to search for highly accurate and efficient quantized models. However, existing QA-NAS approaches, particularly few-bit mixed-precision (FB-MP) methods, do not scale to larger tasks. Consequently, QA-NAS has mostly been limited to low-scale tasks and tiny networks. In this work, we present an approach to enable QA-NAS (INT8 and FB-MP) on large-scale tasks by leveraging the block-wise formulation introduced by block-wise NAS. We demonstrate strong results for the semantic segmentation task on the Cityscapes dataset, finding FB-MP models $33\%$ smaller and INT8 models $17.6\%$ faster than DeepLabV3 (INT8) without compromising task performance.

\end{abstract}





\maketitle

\section{Introduction}

Deep Learning has shown great success in various Computer Vision (CV) tasks, such as object detection \cite{survey_object_detection} and semantic segmentation \cite{survey_semantic_segmentation}, enabling innovative applications like autonomous driving \cite{wu2017squeezedet}. Consequently, a need to efficiently deploy Deep Neural Networks (DNN) on resource-constrained edge devices has arised \cite{singh2023edge}. However, designing accurate and efficient DNNs in terms of, e.g., model size and inference latency, is challenging due to the complexity of the design space. Fortunately, Neural Architecture Search (NAS) \cite{elsken2019neural} has emerged as a solution to automate the design of highly accurate and efficient DNNs. As a result, NAS has recently become the de-facto approach for designing state-of-the-art models for multiple domains, including vision \citep{cai2019once, liu2018darts}, audio \cite{speckhard2023neural}, and radar~\cite{schalk2023radar}.

Traditional NAS needs to train and evaluate each candidate solution stand-alone. Unfortunately, this can restrain its applicability to low-scale tasks and tiny networks, as the cost of training networks stand-alone can rapidly become intractable when dealing with large-scale and compute-intensive tasks. Weight-sharing NAS overcomes this limitation by using a supernet that encompasses all candidate architectures (i.e., the search space) \cite{xie2021weight}. Only the supernet is trained, with subnets of the supernet sharing weights during training, thereby substantially reducing training time. Once the supernet is trained, subnets are sampled and evaluated using subnet performance as a proxy for stand-alone performance to search for the top-performing solutions. However, recent work \cite{chu2021fairnas} has shown that improper or insufficient supernet training in large search spaces can lead to a low correlation between subnet and stand-alone performance, resulting in sub-optimal solutions being found. To address this challenge, Block-wise NAS (BWNAS) \citep{li2020block, moons2021distilling} divides the supernet into several blocks to reduce exponentially the number of subnets in each block compared to the number of subnets in the entire search space. Afterward, subnets are trained block-wisely, allowing them to be better optimized and thereby leading to a higher ranking correlation and better-performing solutions with little compute effort.


While there has been great success in deriving high-performing full-precision architectures, models typically need to be quantized for edge deployment, often leading to accuracy degradations due to the loss of precision \cite{van2023quantization}. However, there is growing evidence that incorporating quantization into the NAS loop, i.e., quantization-aware NAS (QA-NAS), instead of just quantizing models post-search can lead to discovering better-performing quantized models \citep{van2023quantization, van2023bomp}. Furthermore, jointly searching for architectures and few-bit mixed-precision (FB-MP) quantization policies can unlock further efficiency gains compared to homogeneous INT8 quantization, as shown by recent work \citep{wang2020apq, bai2021batchquant}. Unfortunately, existing weight sharing-based QA-NAS methods, particularly FB-MP methods ~\citep{wang2020apq, bai2021batchquant}, do not scale well as these typically present complex and compute-intensive formulations to introduce quantization into the search space. This opens up an opportunity for BWNAS, where due to the effectiveness of its block-wise formulation in segmenting the search space, introducing quantization-awareness into BWNAS could be a suitable approach for scaling weight sharing-based QA-NAS to larger tasks, especially for FB-MP quantization.

In this paper, we present an approach to tackle the challenge of scaling QA-NAS by efficiently introducing quantization awareness into BWNAS. Overall, our contributions are as follows:

\begin{itemize}
    \item We advance the state of the art by introducing quantization-awareness into BWNAS (QA-BWNAS), developing a simple and effective approach to derive optimal INT8 and FB-MP models, achieving strong results on the Cityscapes dataset~\cite{cordts2016cityscapes} for semantic segmentation.
    \item By applying our approach to semantic segmentation, we empirically show that our method is suitable for scaling QA-NAS up to large-scale and compute-intensive tasks, especially for FB-MP quantization.
    \item We present an optimized version of the traversal search algorithm introduced by DNA \cite{li2020block} to reduce the search cost from several hours to only a few seconds.
\end{itemize}

\begin{figure*}[!t]
    \centering
    \includegraphics[width=13cm]{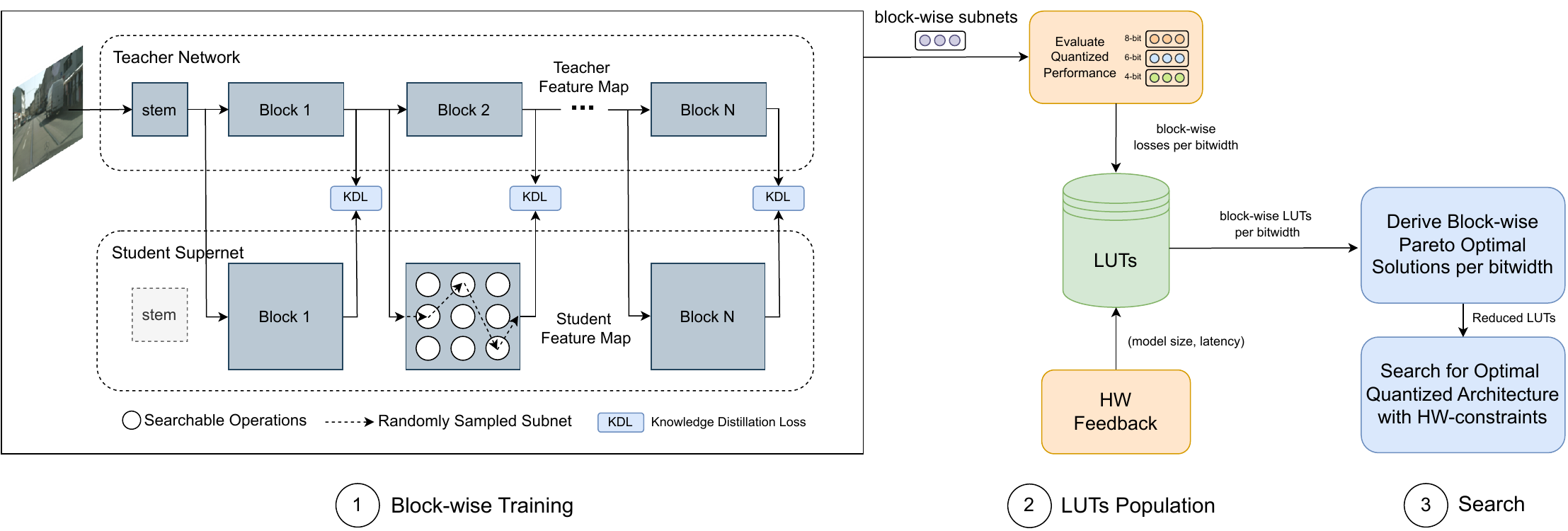}
    \vspace*{-1mm}
    \caption{Overview of our QA-BWNAS approach. (1) We train the blocks in the student supernet via feature-based knowledge distillation. (2) Subnets in each block are then quantized and evaluated in terms of their distillation loss and secondary HW-related metrics (model size and latency) to populate LUTs for searching. (3) We derive the block-wise Pareto optimal subnets per bitwidth to remove the sub-optimal networks from the solution space. Finally, we jointly search for an architecture and quantization policy under a given HW constraint.
    }
    \label{fig:flow}
\end{figure*}

\section{Background and Related Work}

\subsection{Block-wise NAS}
Recent work in weight-sharing NAS indicates that the low-ranking correlation between subnet and stand-alone performance can be attributed to insufficient or improper supernet training in large search spaces \cite{chu2021fairnas}. DNA \cite{li2020block} addresses this by dividing the supernet into blocks (i.e., sub-supernets) to reduce exponentially the search space of candidates in each block compared to the ones in the entire supernet. Afterward, subnets in each block are trained via feature-based knowledge distillation \cite{gou2021knowledge}, using features from a teacher network to supervise the optimization procedure. The block-based optimization facilitates subnets to be better trained, improving the correlation and, thus, leading to better solutions than regular weight-sharing NAS. Finally, the top-performing architecture is found by selecting the subnet on each block that leads to the lowest distillation loss on a validation set.
In contrast, DONNA \cite{moons2021distilling} does not rely on block-level sub-supernets. Instead, it trains block-level subnets directly via knowledge distillation to build a subnet library to fit an accuracy predictor, using the distillation loss as input features. The predictor is then used in conjunction with an evolutionary algorithm to search for high-performing models under constraints.

However, these methods fail to consider quantization as a crucial lever to search for accurate and efficient quantized models. Furthermore, in the case of DONNA, the high compute effort necessary to build the block library makes it an unsuitable approach to search for compute-intensive tasks.

\subsection{Joint Quantization and Neural Architecture Search}
Recent work has explored jointly searching for optimal architectures and quantization policies to derive highly accurate and efficient quantized models.
APQ \cite{wang2020apq} samples a large number of full-precision subnets from a once-for-all (OFA) network \cite{cai2019once} and performs quantization-aware fine-tuning (QAFT) \cite{krishnamoorthi2018quantizing} on these to create a dataset for training a QAFT accuracy predictor. Afterward, the predictor is used to search for a high-performing model given a user-defined HW constraint regarding inference latency or energy. Finally, the best-found model is fine-tuned via QAFT to reach peak performance. In contrast, QFA \cite{bai2021batchquant} introduces a quantizer formulation to train a quantized OFA supernet where quantized models can be directly sampled from the QFA supernet without requiring fine-tuning. Despite the promising results, these methods require thousands of GPU hours to train the networks for searching, which makes them unsuitable for large-scale and compute-intensive tasks.

In this work, we propose an simple yet effective approach to scale QA-NAS and search for optimal quantized DNNs. In contrast with previous work, our method is straightforward and requires little compute, which allows its applicability to higher complexity tasks like semantic segmentation.

\section{Method}

\subsection{Block-wisely supervised training via Knowledge distillation}

Inspired by DNA \cite{li2020block}, we define and divide a supernet into $N$ blocks, using the features of a state-of-the-art NN model (i.e., the teacher network) to supervise the block-wise training procedure. Specifically, the output feature map of the ~$(i-1)$-th block of the teacher is used as the input of the ~$i$-th block of the student supernet, with the output features of the teacher's ~$i$-th block serving as supervision, as shown in Fig.~\ref{fig:flow}~(1). During block-wise training, subnets are sampled randomly and optimized to minimize the per-channel noise-to-signal-power ratio (NSR) \cite{moons2021distilling}:

\begin{equation}
    \mathcal{L}_{NSR}(\mathcal{Y}_n, \hat{\mathcal{Y}}_{n}) = \frac{1}{C} \sum_{c=0}^C \frac{\| \mathcal{Y}_{n,c} - \hat{\mathcal{Y}}_{n,c} \|^2}{\sigma^2_{n,c}}
\label{eq:nsr}
\end{equation}
where $\mathcal{Y}_n$ is the teacher's target feature map of block $B_n$, $\hat{\mathcal{Y}}_{n}$ is the output feature map of the candidate subnet under optimization, and $\sigma^2_{n,c}$ is the variance of $\mathcal{Y}_{n,c}$. Furthermore, $C$ is the number of channels in the feature map.
By optimizing this objective function, subnets are trained to closely mimic the teacher network features.

\subsection{Quantization-Aware Block-Wise NAS} \label{sec: method-3}

Having trained each block, we apply post-training quantization (PTQ) \citep{krishnamoorthi2018quantizing} on each candidate subnet, using the weights learned during block-wise training and evaluate the distillation loss of the quantized subnets to assess the performance of these low-precision architectures. As shown in Fig. \ref{fig:flow} (2), LUTs are populated with the quantized loss and then used to search for the optimal quantized model. In contrast with QAFT, we opt for PTQ, as quantizing and evaluating the subnets introduces less compute effort since no fine-tuning is performed after quantization.

While INT8 quantization can be an excellent way to improve the efficiency of NN models, it has been shown that DNNs can be further compressed without introducing considerable performance degradation by selectively quantizing specific layers to fewer-bit representations (i.e., below 8-bit), leading to FB-MP models with a smaller footprint and improved real-time performance on HW with on-board support. However, the biggest challenge resides in deriving the FB-MP quantization policy that maps bitwidths to each layer, determining which layers should be quantized further and which should remain in higher precision.

We introduce FB-MP into our QA-BWNAS by quantizing the subnets of each block to fewer-bit representations, creating a total of $N \times B$ LUTs, where $N$ is the number of blocks, and $B$ is the number of bitwidths in the search space. These LUTs are then concatenated to jointly search for an optimal architecture and FB-MP policy. 

For secondary HW-related constraints, we populate additional LUTs containing the contribution of each subnet in terms of model size and inference latency. These LUTs are then used for searching for optimal models under constraints.

\subsection{Searching within Pareto optimal solutions on each block.}

One challenge, however, is that introducing LUTs for each target bitwidth can substantially increase searching time, as the search space of each block increases linearly as a function of the bitwidth choices in the search space. Therefore, we present an optimization of DNA's search algorithm to rapidly search for optimal quantized models.

DNA's traversal search algorithm first ranks the subnets in each block. If we only wish to find the most accurate solution, the search is reduced to simply picking the top rank for each block in terms of its NSR loss. However, when dealing with secondary constraints, DNA subtly visits all models in the search space. This can be highly inefficient as millions of sub-optimal (i.e., non-Pareto optimal) models are explored. Therefore, to reduce searching time, we first derive the Pareto optimal solutions in terms of the NSR loss and a secondary HW-related metric for each LUT. Then, we remove the sub-optimal, non-Pareto Optimal solutions. Consequently, searching through the LUTs is reduced to a few seconds. Afterward, we search within these reduced LUTs to find the top-performing models under the user-defined constraints.

Finally, the found architecture is retrained to achieve peak performance and then quantized for edge deployment using the quantization policy derived during the search.

\section{Experimental Setup} \label{sec:exp}

\textbf{Dataset and Teacher Network.} We evaluate our method on the Cityscapes dataset \cite{cordts2016cityscapes}. We only use fine-grain annotations for all experiments and an input resolution of 512 $\times$ 1024 for all trainings. For the teacher network, we use DeeplabV3 \cite{chen2017rethinking} with a MobileNetV2 \cite{sandler2018mobilenetv2} encoder and output stride of 8. Note that, as we target edge devices, we do not use the ASPP module.

\textbf{Search Space.} We only search for the encoder component of the semantic segmentation network and search for MBConv layers with kernel sizes $\{3, 5, 7\}$ and expansion ratios $\{3, 6\}$ for each block, as shown in Table \ref{table:block_details}.

\textbf{Block-wise Training.} We initialize each block with ImageNet \cite{deng2009imagenet} weights via weight remapping \cite{fang2020fna} to circumvent the costly pre-training process. Further, each block is trained for 13333 iterations, for an overall total of 80K iterations, which is the total number of iterations required to train a single model. Empirically, we find that this is an adequate training scheme to find good-performing models. Further, we use $[0.002, 0.005, 0.005, 0.005, 0.005, 0.002]$ as the learning rate for each block using SGD with a momentum of 0.9.

\begin{table}[t]
\caption{Search space details for each block. "L\#" and "CH\#" represent the number of layers and channels of each block.}
\label{table:block_details}
\centering
\begin{tabular}{c|cc|cc}
\hline
 & \multicolumn{2}{c|}{Teacher} & \multicolumn{2}{c}{Student Supernet} \\ \hline
Block & L\#          & CH\#          & L\#              & CH\#              \\ \hline
1     & 2            & 24            & 3                & 24                \\ \hline
2     & 3            & 32            & 3                & 32                \\ \hline
3     & 4            & 64            & 4                & 64                \\ \hline
4     & 3            & 96            & 4                & 96                \\ \hline
5     & 3            & 160           & 3                & 160               \\ \hline
6     & 1            & 320           & 1                & 320               \\ \hline
\end{tabular}
\end{table} 

\textbf{Quantization and LUT Population Details.} We search for $\{4, 6, 8\}$-bit weight subnets on each block. All activations and weights of the not searchable components of the model are quantized to 8 bits. All models are quantized using min-max per-channel PTQ via QKeras \cite{Coelho_2021}. The NSR of each subnet is evaluated using 20 samples from the Cityscapes validation set. For HW-related secondary metrics, we calculate model size theoretically based on the bitwidth and number of parameters of each layer. Inference latency is profiled directly on an i.MX8M Plus using the TFLite inference engine.

\textbf{Searching.} To derive a dense Pareto front, we search for multiple models across various constraints in terms of model size and inference latency. Note that, for latency, we only search for INT8 models, as the i.MX8M Plus does not have on-board support for accelerating FB-MP models.

\textbf{Retraining.} Having derived optimal architectures, we initialize these with ImageNet weights via weight remapping and train them for 80K iterations, following the same training settings as the DeepLabV3 teacher network \cite{chen2017rethinking}. Finally, models are quantized via PTQ using the quantization policy derived during the search. Each model training takes approximately 5 GPU hours. All experiments are performed on a single NVIDIA RTX 8000 GPU.

\begin{figure}[t]
    \centering
    \includegraphics[width=7.0cm]{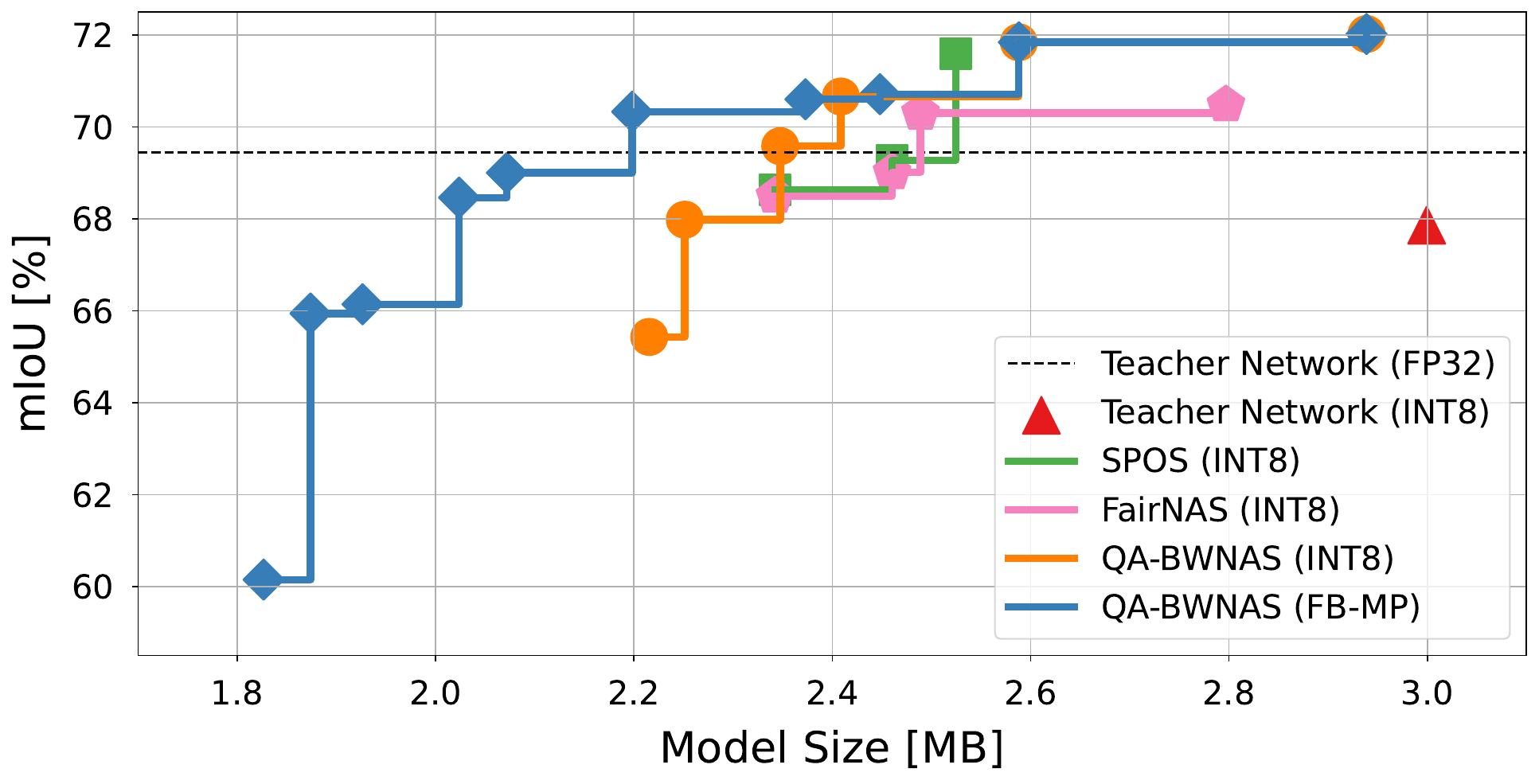}
    \vspace*{-1mm}
    \caption{QA-BWNAS derives highly optimized solutions, reducing model size up $25\%$ (INT8) and $33\%$ (FB-MP), while retaining mIoU on the Cityscapes validation set.}
    \label{fig:qabwnas_int8}
\end{figure}

\begin{table}[t]
\vspace*{-1.5mm}
\caption{Compute cost (GPU Hours) comparison between QA-BWNAS and state-of-the-art weight sharing methods extended towards QA-NAS. \textpm~indicates the search is performed without removing the non-Pareto optimal solutions from the LUTs. $N$ indicates number of searched models.}
\label{table:compute_effort}
\centering
\begin{tabular}{ccccc}
\hline
\textbf{Method}  & \textbf{Train} & \textbf{\begin{tabular}[c]{@{}c@{}}LUT \\ Population\end{tabular}} & \textbf{Search}  \\ \hline
QA-BWNAS (FP-MP)\textpm & 4.05          & 44.61                                                          & 14 $\times$ N                  \\  \hline
QA-BWNAS (FP-MP) & 4.05          & 44.61                                                          & 0                  \\
QA-BWNAS (INT8)  & 4.05          & 14.87                                                          & 0                  \\
FairNAS (INT8)   & 3.5           & -                                                                  & 7.5                     \\
SPOS (INT8)      & 4.5           & -                                                                  & 7.5                     \\ \hline
\end{tabular}
\end{table}

\section{Results}

\subsection{Homogeneous INT8 Quantization}

As shown in Fig. \ref{fig:qabwnas_int8}, QA-BWNAS (INT8) derives a Pareto front of solutions, substantially outperforming the teacher network. In particular, we find a model that achieves $72.02\%$ mean Intersection over Union (mIoU), representing approximately a $4.2$ percentage points (pp.) increase over the quantized teacher's performance while slightly reducing model size, even outperforming the full-precision teacher. Remarkably, we also find a model approximately $25\%$ smaller, with slightly higher accuracy as the INT8 quantized teacher network. These results are interesting since intuitively, one may think that a student network would not be able to surpass the teacher's performance. However, we empirically find that smaller students can match or surpass the teacher. We hypothesize that this is due to regularization effects. Moreover, the effectiveness of our optimized search algorithm to reduce search time from hours to just a few seconds per model can be observed in Table~\ref{table:compute_effort}.

For comparison with our method, we select two existing state-of-the-art weight-sharing NAS methods, namely, SPOS \cite{guo2020single} and FairNAS \cite{chu2021fairnas}, and extend them to search for optimal INT8 semantic segmentation models. These methods are selected for comparison as these require relatively little compute effort, which makes them suitable candidates for searching on the semantic segmentation task. The search spaces are the same as QA-BWNAS to ensure a fair comparison. Moreover, supernets are trained for 80K iterations to ensure equivalent training as our block-wise setting. Having trained both supernets, we randomly sample 100 architectures, quantize them and select the Pareto optimal subnets, re-training them stand-alone to achieve peak performance. As shown in Fig. \ref{fig:qabwnas_int8}, QA-BWNAS (INT8) outperforms both SPOS and FairNAS, achieving a better trade-off with little additional compute, as shown in Table~\ref{table:compute_effort}.

\begin{figure}[t]
    \centering
    \includegraphics[width=7.0cm]{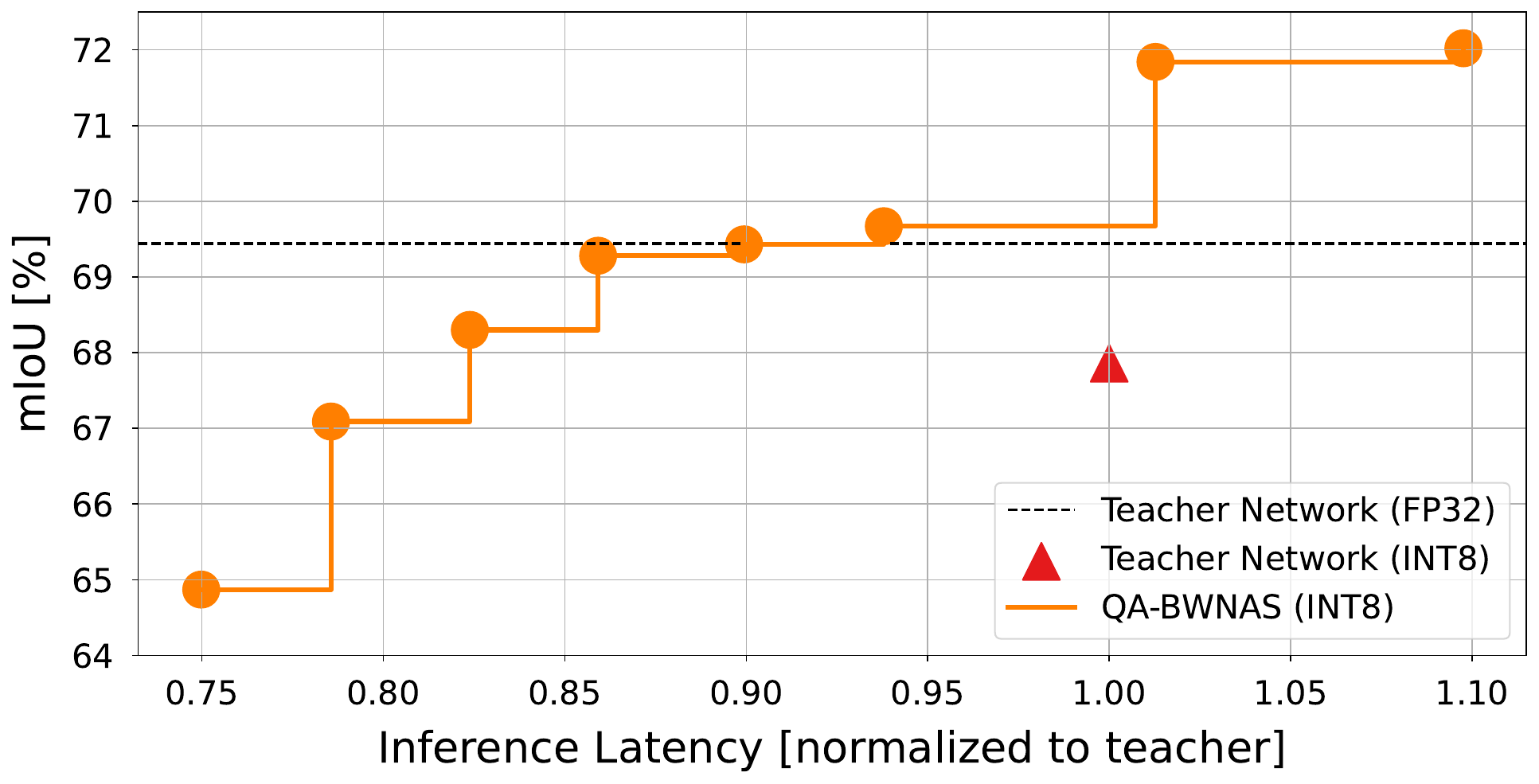}
    \vspace*{-1mm}
    \caption{QA-BWNAS reduces inference latency on an i.MX8M Plus up to $17.6\%$ while retaining mIoU performance on the Cityscapes validation set.}
    \label{fig:qabwnas_int8_lat}
\end{figure}

Furthermore, the results of searching for optimal INT8 models across various inference latency constraints are presented in Fig. \ref{fig:qabwnas_int8_lat}. As shown, QA-BWNAS (INT8) finds models that achieve a $17.6\%$ inference latency reduction on the i.MX8M Plus while slightly outperforming the teacher model in terms of mIoU on the cityscapes dataset. By optimizing models for latency and model size, we show that our approach can accommodate various secondary objectives.

\subsection{Few-bit Mixed-precision Quantization}

As shown in Fig. \ref{fig:qabwnas_int8}, QA-BWNAS (FP-MP) exhibits a Pareto dominance over the INT8 solutions, emphasizing the benefits of jointly searching for the optimal architecture and FB-MP policy. Remarkably, we find a model that achieves 0.6 pp. higher mIoU with a model size reduction of approximately $33\%$ compared to the teacher network. Regarding compute effort, FB-MP search requires a total of 48.66 GPU Hours. This is expected, as the FB-MP search requires populating extra LUTs per block for each searchable bit-width.

\section{Conclusion}
This work presents QA-BWNAS, a simple yet effective approach to scale QA-NAS by introducing quantization awareness into BWNAS. By demonstrating strong results on the semantic segmentation task, we show that our method presents a suitable alternative to scale QA-NAS towards large-scale and compute-intensive tasks. Moreover, we introduce an optimization of the traversal search algorithm presented in DNA \cite{li2020block}, which allows us to reduce search time from hours to just a few seconds. 

\section{Acknowledgements}

This work was supported by Key Digital Technologies Joint Undertaking (KDT JU) in EdgeAI “Edge AI Technologies for Optimised Performance Embedded Processing” project, grant agreement No 101097300.

\bibliographystyle{ACM-Reference-Format}
\bibliography{references}

\end{document}